\def\year{2022}\relax
\documentclass[letterpaper]{article} 
\usepackage{aaai22}  
\usepackage{times}  
\usepackage{helvet}  
\usepackage{courier}  
\usepackage[hyphens]{url}  
\usepackage{graphicx} 
\usepackage{natbib}  
\usepackage{caption} 
\DeclareCaptionStyle{ruled}{labelfont=normalfont,labelsep=colon,strut=off} 
\usepackage{algorithm}
\usepackage{algorithmic}

\usepackage{times}
\usepackage{epsfig}
\usepackage{graphicx}
\usepackage{amsmath}
\usepackage{amssymb}
\usepackage{threeparttable}
\usepackage{lscape}
\usepackage{array}
\usepackage{booktabs}
\usepackage{caption}
\usepackage{subcaption}
\usepackage{bm}
\usepackage{multirow}

\usepackage{hyperref}

\usepackage{newfloat}
\usepackage{listings}
\floatstyle{ruled}
\newfloat{listing}{tb}{lst}{}
\floatname{listing}{Listing}
\title{AAAI Press Formatting Instructions \\for Authors Using \LaTeX{} --- A Guide}

\title{Commonsense Knowledge from Scene Graphs for Textual Environments}
\author {
    Tsunehiko Tanaka, \textsuperscript{\rm 1,2}\thanks{This work was done during an internship at IBM Research.} \:
    Daiki Kimura, \textsuperscript{\rm 1} \:
    Michiaki Tatsubori \textsuperscript{\rm 1}
}
\affiliations {
    \textsuperscript{\rm 1} IBM Research \:\:\:
    \textsuperscript{\rm 2} Waseda University\\
    tsunehiko@fuji.waseda.jp, daiki@jp.ibm.com, mich@jp.ibm.com
}

\usepackage{bibentry}

\begin{document}

\maketitle

\begin{abstract}
Text-based games are becoming commonly used in reinforcement learning as real-world simulation environments. They are usually imperfect information games, and their interactions are only in the textual modality. To challenge these games, it is effective to complement the missing information by providing knowledge outside the game, such as human common sense. However, such knowledge has only been available from textual information in previous works. In this paper, we investigate the advantage of employing commonsense reasoning obtained from visual datasets such as scene graph datasets. In general, images convey more comprehensive information compared with text for humans. This property enables to extract commonsense relationship knowledge more useful for acting effectively in a game. We compare the statistics of spatial relationships available in Visual Genome (a scene graph dataset) and ConceptNet (a text-based knowledge) to analyze the effectiveness of introducing scene graph datasets. We also conducted experiments on a text-based game task that requires commonsense reasoning. Our experimental results demonstrated that our proposed methods have higher and competitive performance than existing state-of-the-art methods.
\end{abstract}

\section{Introduction}
\begin{figure}[t]
    \centering
    \includegraphics[width=\linewidth]{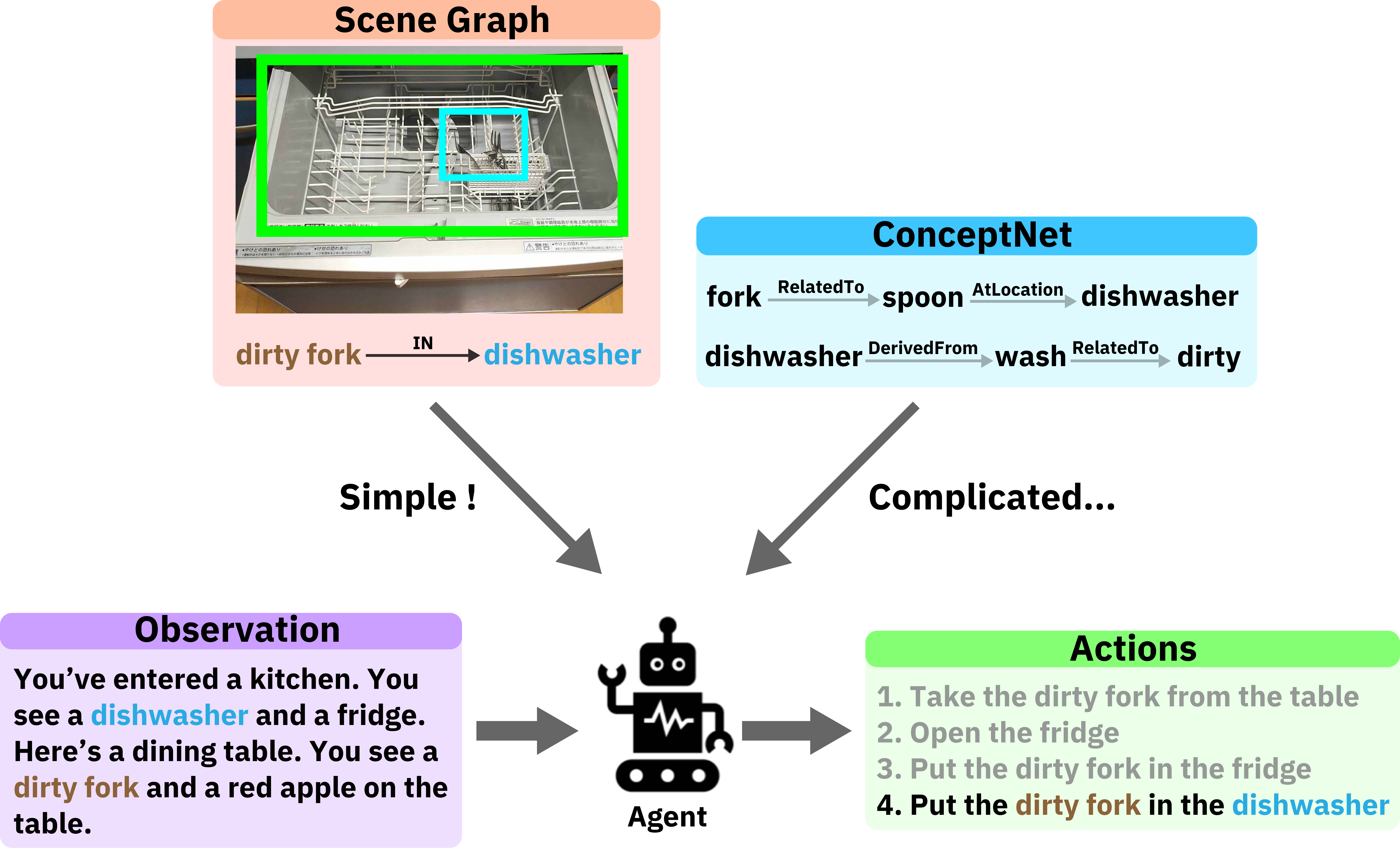}
    \caption{Illustration of our commonsense acquisition from scene graphs. To provide commonsense: $\textrm{dirty fork} \rightarrow \textrm{IN} \rightarrow \textrm{dishwasher}$ to an agent, a single image is sufficient for scene graphs (top left), but ConceptNet requires several graphs to be combined, which is redundant.}
    \label{fig:teaser}
 \end{figure}

Reinforcement learning (RL) is a type of machine learning method that has a great advantage of not requiring labeled data and has been used in various simulation environments~\cite{dqn,alphago,daqn,marioirl}. Since textual conversation agents are commonly used in our daily lives in the real world, text-based environments, where both observation and action spaces are restricted to the modality of text, have been attracting attention. RL in such environments requires developing an agent to have language comprehension skills by natural language process and sequential decision-making in the complex environment. This means the textual observation contains a lot of noisy information and the problem of partial observability.

Text-based games are a partially observable Markov decision process (POMDP)~\cite{pomdp} where the agent cannot observe the entire information from the text given by the environment. TextWorld~\cite{cote18textworld} is a textual game generator and extensible sandbox learning environment for RL agents, and various methods have been proposed for this game to compensate for the missing information~\cite{kimura2020reinforcement,murugesan2021textbased,carta-etal-2020-vizhints,murugesan2021eye,ALFRED20,kimura-etal-2021-loa,kimura-etal-2021-neuro,chaudhury-etal-2021-neuro}. There are three types of extensions: external knowledge, new modality, and logical rule extraction.
External knowledge that is useful for training agents from humans or other domain sources. A study reports commonsense knowledge is an important aspect of human intelligence~\cite{murugesan2021textbased}. In this study, TextWorld Commonsense~(TWC), which requires commonsense as external knowledge, is proposed as an extension of TextWorld. The task of the TWC game is cleaning up a room, and the commonsense in this game is mainly place information for each object. The same study also includes a baseline agent for TWC games that uses a commonsense subgraph extracted from external knowledge (we call this model \textit{TWC agent} and the environment \textit{TWC games} to distinguish them). Another study reported that introducing external knowledge from humans as logical functions helps the training of the agent~\cite{kimura2020reinforcement}. 
New modality information extracted from observations or action text can be introduced to make decisions~\cite{carta-etal-2020-vizhints, murugesan2021eye, ALFRED20}. In these methods, visual information from images or videos is commonly used since it has been used in many other studies~\cite{Tanaka_2021_CVPR,9093428} to understand attention and sequential information in decision making. 
Logical rule extraction can be exploited to improve the speed of training and interpretability of the agent~\cite{kimura-etal-2021-neuro,chaudhury-etal-2021-neuro}. Since commonsense knowledge is normally represented by a graph structure, the logical rule representation is compatible with commonsense knowledge. 

However, at the time of writing, there has been no research that utilizes the benefits of these multiple extensions to compensate for missing information. In particular, we hypothesize that the commonsense knowledge of object place relationships that are used in TWC games can be easily obtained from visual information. For example, instead of stating the place name of each object, operators can display a picture of a tidy room, which is a quicker explanation for humans.

In this paper, we propose a novel agent that challenges a TWC game by leveraging visual scene graph datasets to obtain commonsense. The original TWC agent~\cite{murugesan2021textbased} constructs a commonsense subgraph from ConceptNet~\cite{ConceptNet2017}, which is textual knowledge, but it is necessary to combine many graphs to obtain one commonsense and to create a complicated subgraph. In fact, Murugesan~et~al. prepared a `manual' commonsense subgraph from ConceptNet to tackle this complexity of graphs in their study. However, since scene graph recognition achieves high accuracy from complex images, visual information can deliver various detailed and organized graph information all at once. Figure~\ref{fig:teaser} shows an example for the acquisition of commonsense knowledge from scene graphs in an image. In this example, despite ConceptNet having redundant information for extracting a commonsense subgraph, the proposed extraction from scene graphs has necessary and sufficient information for the cleaning-up task. Furthermore, relationships from scene graphs also contain direct spatial relationships such as ``on'' or ``in'' (Figure \ref{fig:relationships}) between objects because agents need to determine an object's place in the TWC game. Therefore, we use scene graph datasets as visual external knowledge. A scene graph dataset contains a large number of graphs that represent the relationships between entities in images. We use Visual Genome (VG)~\cite{krishna2017visual} as a scene graph dataset, which is the most commonly used, and compare its statistics with ConceptNet. We also conduct experiments to evaluate the performance of agents with commonsense knowledge from a scene graph dataset in RL on text-based games.

\section{Related Work}
\subsection{Text-based RL Games} 
Text-based interactive RL games has been gaining the focus of many researchers due to the development of environments such as TextWorld~\cite{cote18textworld} and Jericho~\cite{hausknecht2019interactive}. In these games, RL agents are required to understand the high-level context information from only textual observation. To overcome this difficulty, a number of prior works on these environments have extracted new information from textual observations: knowledge graphs, visual information, and logical rule.

Knowledge graphs represent relationships between entities like real-world objects and events, or abstract concepts. A new text-based environment, called ``TextWorld Commonsense'', was proposed in \cite{murugesan2021textbased} to infuse RL agents with commonsense knowledge and developed baseline agents using a commonsense subgraph constructed from ConceptNet\cite{Liu2004ConceptNetA, ConceptNet2017} as an external knowledge. We use this work as a baseline method, and introduce a new type of commonsense from visual datasets. Worldformer~\cite{ammanabrolu2021learning} represents environment status as a knowledge graph and uses a world model to predict changes caused by an agent's actions and generates a set of contextually relevant actions.

While knowledge graphs are useful for organizing abstract information from only text descriptions, visual information enables the agent to obtain a detailed locational situation like human imagination and visualization. The most important issue in using visual information is how to obtain it from only textual observation in text-based games. VisualHints~\cite{carta-etal-2020-vizhints} proposed an environment that can automatically generate various hints about game states from textual observation and changes the difficulty level depending on their type. The main sources of images in~\cite{murugesan2021eye} are retrieved from the Internet and generated from a text-to-image pre-trained model, AttnGAN~\cite{Tao18attngan} with given text descriptions. ALFWorld~\cite{ALFWorld20} combines TextWorld and an embodied simulator called ALFRED~\cite{ALFRED20} to obtain information on two modalities. Shridhar~et~al. proposed an agent that first learns to solve abstract tasks in TextWorld, then transfers the learned high-level policies to low-level embodied tasks in ALFRED.

In addition, even if we use the aforementioned methods, improvements in the speed of training are few and the interpretability of the trained network is still missing. A number of studies~\cite{kimura-etal-2021-neuro,chaudhury-etal-2021-neuro} proposed novel approaches to extract symbolic first-order logics from text observations, and select actions by using neuro-symbolic Logical Neural Networks~\cite{riegel2020logical}. These logical representations are compatible with commonsense graph structures.

As previously described, there have been various approaches using knowledge graphs, visual information, and logical rules. However, at the time of writing, there has been no method that combines any of them. Therefore, we propose an approach to extract and utilize knowledge graphs from visual information.

\subsection{Scene Graph Dataset}
A scene graph is a structured representation of the relationships between objects in a scene. To train a scene graph generation model, a number of datasets have been created. VG~\cite{krishna2017visual} is a large-scale scene graph dataset that is most commonly used these days because it contains various elements such as objects, attributes, relationships, QA descriptions, and so on. Since scene graphs can provide a large number of visual relationships in a single image, we use VG datasets as external knowledge for training agents.

\section{ConceptNet vs Scene Graph Datasets}
In this section, to show scene graph datasets are effective as external knowledge for solving TWC games, we compare ConceptNet and scene graph datasets. We first show the statistics of ConceptNet, VG~\cite{krishna2017visual}, and manual commonsense knowledge designed in \cite{murugesan2021textbased}. Next, we compare ConceptNet and VG in terms of similarity to the manual commonsense knowledge. VG is the most commonly used scene graph dataset. The manual commonsense knowledge is manually extracted from ConceptNet to include only the pairs of an object in TWC games and goal location for each object. Since the entities are directly related to actions in the games, the agent with this manually-crafted information is more effective for solving the games. Therefore, external knowledge that is similar to the manual commonsense knowledge are comfortable with this task.

\subsection{Knowledge Statistics}

\begin{table}
   \centering
   \begin{tabular}[h]{>{\centering}m{0.27\linewidth}>{\centering}m{0.15\linewidth}>{\centering}m{0.2\linewidth}>{\centering\arraybackslash}m{0.15\linewidth}}
       \toprule
        & entity & relationship & triplet \\
       \midrule
       VG & 63,686 & 36,550 & 662,934 \\
       ConceptNet & 38,556 & 46 & 298,394 \\
       Manual & 111 & 2 & 132 \\
       \bottomrule
   \end{tabular}
   \caption{Statistics of the three types of external knowledge. Each item denotes the number of species. Manual is manual commonsense knowledge used in \cite{murugesan2021textbased}.}
   \label{tab:statistics}
\end{table}

\begin{figure*}[h]
   \begin{center}
      \begin{tabular}{c}
         \begin{minipage}{0.5\linewidth}
            \centering
            \includegraphics[width=\linewidth]{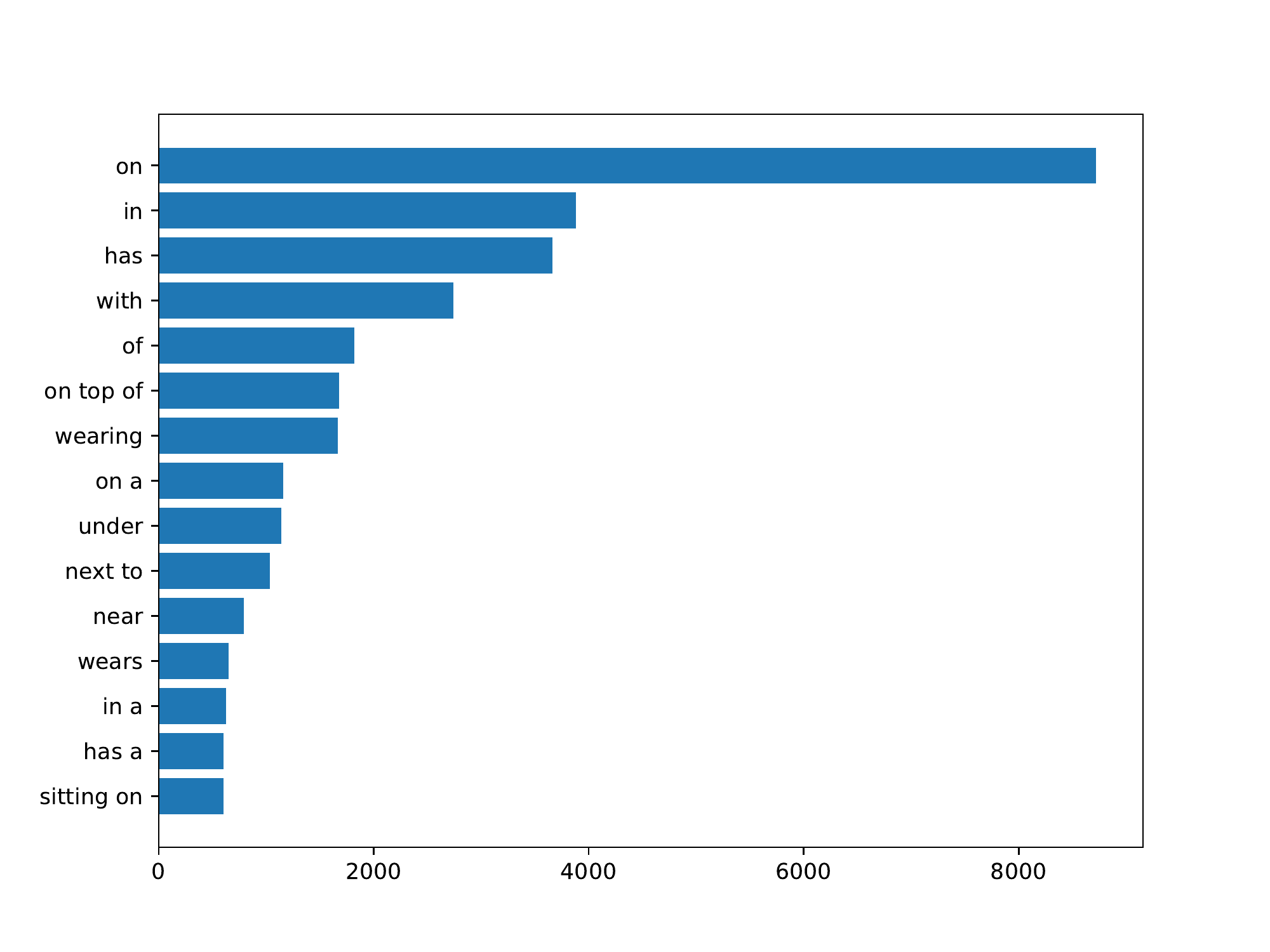}
            \subcaption{VG}
            \label{fig:vg_relationships}
         \end{minipage}
         \begin{minipage}{0.5\linewidth}
           \centering
           \includegraphics[width=\linewidth]{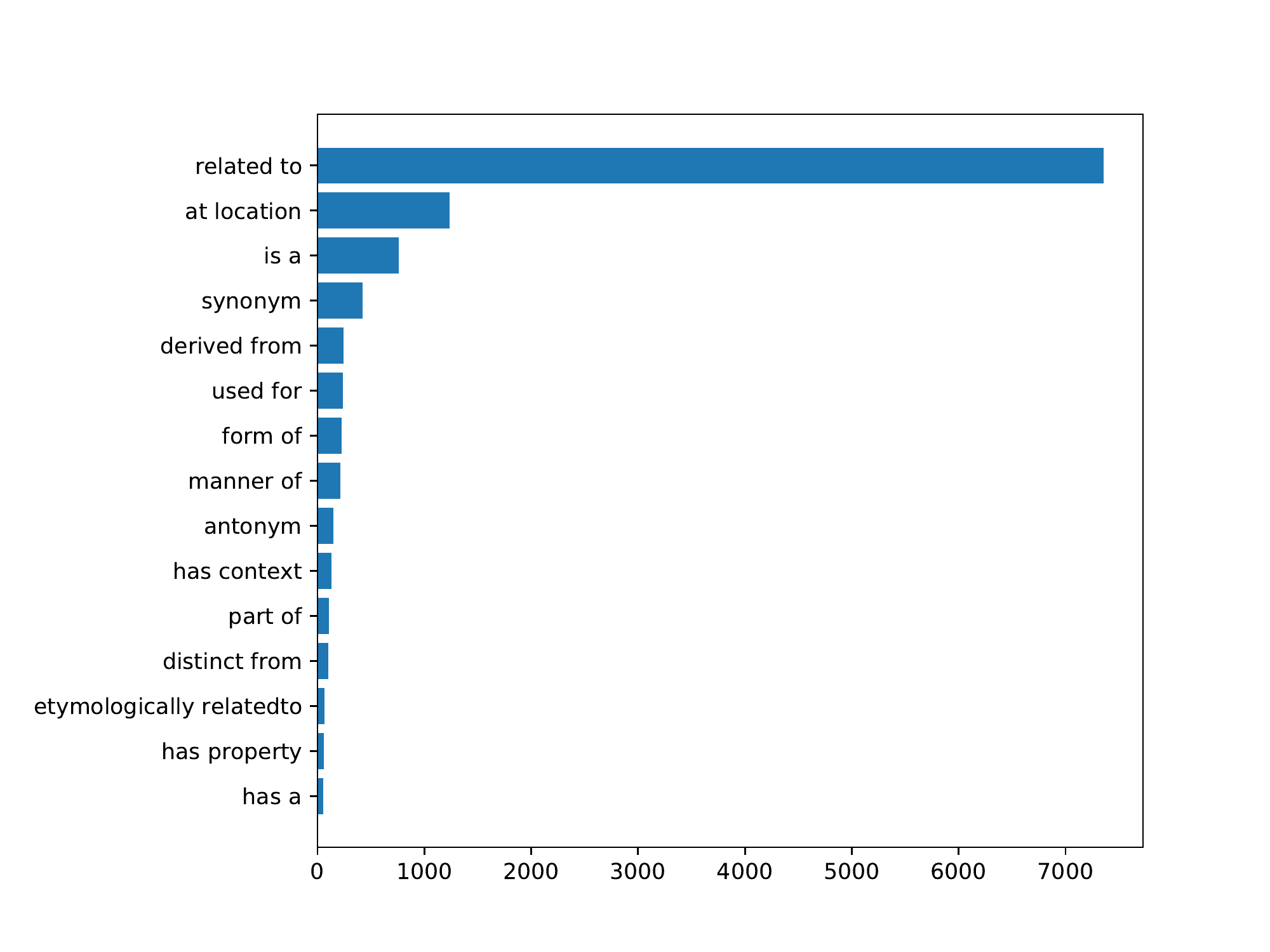}
           \subcaption{ConceptNet}
           \label{fig:cn_relationships}
        \end{minipage}
      \end{tabular}
      \caption{Histogram of relationships (top 15) included in VG and ConceptNet. VG has much more spatial relationships than ConceptNet.}
      \label{fig:relationships}
   \end{center}
\end{figure*}

We summarize the statistics of the three types of external knowledge in Table~\ref{tab:statistics}. In general for all external knowledge, each graph is represented as a triplet $\langle e_1, r_{12}, e_2 \rangle$: $e_1, e_2$ denote entities in an image, and $r_{12}$ denotes a relationship between $e_1$ and $e_2$. In Table~\ref{tab:statistics}, `entity', `relationship', and `triplet' indicate the number of species of $e$, $r$, $\langle e_1, r_{12}, e_2 \rangle$, respectively. The huge difference between ConceptNet and VG is the number of species of relationships, and this indicates that VG has more detailed information in relationships than ConceptNet. Figure~\ref{fig:relationships} shows an example of $r$ in these datasets. In ConceptNet, the spatial relationship is only `at location', which is the second most common, but its ratio to the total is low. In contrast, spatial relationships such as `on', `in', and `under' dominate VG. In addition, `has' and `with' can also express spatial relationships, such as $\langle\mathrm{building}, \mathrm{has}, \mathrm{window} \rangle$ and $\langle \mathrm{window}, \mathrm{with}, \mathrm{building} \rangle$. Thus, we can see that VG has a larger number of spatial relationships than ConceptNet. However, the manual commonsense knowledge has two species of relationships: 186~`at location' and 6~`related to'. This means that spatial relationships are important for solving TWC games, and VG has an advantage in this respect.

\subsection{Similarity to Manual Commonsense Knowledge}
To examine whether VG contains knowledge graphs useful for solving TWC games, we compare ConceptNet and VG in terms of similarity to the manual commonsense knowledge. We calculate the similarities for both entity $e$ and entity pairs $\{e_1, e_2\}$ as described in the following.

\subsubsection{Entity $e$}
Given an entity $e^V_i$ from VG, we use GloVe~\cite{pennington-etal-2014-glove} embeddings to represent $e^V_i$ as a $d$-dimensional vector $\bm{z}^V_i$, where $\bm{z}^V_i \in \mathbb{R}^d$
is the word embedding of the entity. Similarly, the embedding of each entity in ConceptNet $e^C_j$ and in manual commonsense knowledge $e^M_k$ are denoted as $\bm{z}^C_j, \bm{z}^M_k$, respectively. We calculate the similarity $s^{eV}_ik$ between an entity in VG $e^V_i$ and in manual commonsense knowledge $e^M_k$ by Eq. ~\ref{eq:entity_sim}.
\begin{equation}
   s^{eV}_{ik} = \mathrm{cos\_similarity}(\bm{z}^V_i, \bm{z}^M_k)
   \label{eq:entity_sim}
\end{equation}
Similarly, Eq.~\ref{eq:entity_sim} is executed for entities in ConceptNet. We count the number of entities whose similarity is above the threshold~$0.7$.

\subsubsection{Pair of entities $\{e_1, e_2\}$}
We also compare sets of pairs of $e_1$ and $e_2$ from these datasets in terms of similarity to the 132~pairs in the manual commonsense knowledge. In the same way as the entities previously described, we use GloVe to represent $e^V_{i1}$ and $e^V_{i2}$ from triplet $t^V_i = \langle e^V_{i1}, r^V_{i1i2}, e^V_{i2} \rangle$ in VG as $d$-dimensional vectors $\bm{z}^V_{i1}, \bm{z}^V_{i2}$, respectively. Similarly, $\bm{z}^C_{j1}, \bm{z}^C_{j2}$ and $\bm{z}^M_{k1}, \bm{z}^M_{k2}$ are the embeddings of entities from each triplet in ConceptNet $t^C_j = \langle e^C_{j1}, r^C_{j1j2}, e^C_{j2} \rangle$ and manual commonsense knowledge $t^M_k = \langle e^M_{k1}, r^M_{k1k2}, e^M_{k2} \rangle$, respectively. We calculate the similarity using the sum of both embeddings of entities in a triplet. Thus, the similarity $s^{pV}_{ik}$ is given by Eq.~\ref{eq:relationship_similarity}.
\begin{equation}
   s^{pV}_{ik} = \mathrm{cos\_similarity}(\bm{z}^V_{i1}+\bm{z}^V_{i2}, \bm{z}^M_{k1}+\bm{z}^M_{k2})
   \label{eq:relationship_similarity}
\end{equation}
For ConceptNet, we use Eq.~\ref{eq:relationship_similarity} similarity. We set a threshold to~$0.65$ and count pairs over the threshold. 

\begin{table}
   \centering
   \begin{tabular}[h]{>{\centering}m{0.3\linewidth}>{\centering}m{0.2\linewidth}>{\centering\arraybackslash}m{0.2\linewidth}}
       \toprule
        & entity & pair \\
       \midrule
       VG & 147 & 58,005 \\
       ConceptNet & 156 & 11,550\\
       \bottomrule
   \end{tabular}
   \caption{Comparison of the number of entities and pairs similar to manual commonsense knowledge in external knowledge. A pair is a combination of $e_1$ and $e_2$ from a triplet $\langle e_1, r_{12}, e_2 \rangle$. The number of entities is not very different, but the number of pairs is much higher in VG than in ConceptNet.}
   \label{tab:comparison}
\end{table}

The results of the entity and pair counts are summarized in Table \ref{tab:comparison}. Although there is no significant difference in the number of types of entities, VG has more pairs associated with the manual commonsense knowledge, which indicates that VG has more game-related relationships than ConceptNet.

The aforementioned comparisons show that TWC games need more spatial relationships in the external knowledge, and VG is more effective for solving TWC games than ConceptNet.

\section{Proposed Method}
\subsection{Previous TWC agent}
The proposed methods extend the TWC agent~\cite{murugesan2021textbased}, which is a baseline model for TextWorld Commonsense. We briefly explain the network architecture as follows.

The TWC agent consists of the six components: (a)~\textit{action encoder}, which encodes all admissible actions~$a$, (b)~\textit{observation encoder}, which encodes the observation~$o_t$, (c)~\textit{context encoder}, which encodes the dynamic context~$C_t$, (d)~\textit{dynamic commonsense subgraph}, which is commonsense information~$G_C^t$ extracted by the agent, (e)~\textit{knowledge integration}, which combines the information from textual observation and the extracted commonsense subgraph, and (f)~\textit{action selection}, which selects an action from given action candidates. We subsequently describe \textit{dynamic commonsense subgraph} and \textit{knowledge integration}, which are important for this paper.

For \textit{dynamic commonsense subgraph}, the TWC agent retrieves commonsense from external knowledge like ConceptNet\cite{ConceptNet2017} and updates a subgraph by combining it with the graph at a previous time step. At time~$t$, the agent first extracts entities involving game status from textual observation and then obtains a set of cumulative entities~$E_t$ by combining it with the entities from the previous graph $G_C^{t-1}$. The commonsense subgraph $G_C^t$ is constructed automatically from~$E_t$ and \textit{Context Direct Connections (CDC)}, which is another algorithm of external knowledge. For CDC, the entities are split into two groups in accordance with their attributes, and then links between the groups are added.

For \textit{knowledge integration}, the TWC agent encodes the commonsense subgraph and integrates the graph embedding vector with the observation context feature. In the encoding phase, the node embedding is first extracted from the commonsense subgraph using a pre-trained knowledge graph embedding called Numberbatch~\cite{numberbatch} and a \textit{sentinel} vector~\cite{lu2017knowing} is added to enable the attention to not attend to any specific nodes in the commonsense subgraph. These embeddings are updated by messages passing between the nodes of graph attention networks~(GAT)~\cite{velickovic2018graph}. In the integration phase, \textit{Co-Attention} is used, which is a bidirectional attention flow layer between the observational context and the commonsense subgraph.

TWC agent is an attractive design for RL agents on text-based games, and it accesses a commonsense and uses it while selecting actions. However, the source format of external knowledge is limited to text. Since textual knowledge such as ConceptNet is very useful, it is redundant because multiple concepts need to be concatenated to express more detailed information. Therefore, they prepared manually retrieved graphs in the paper~\cite{murugesan2021textbased}. The manual graphs contain direct connections for the objects and their goal locations.

\subsection{Proposed Method}
From the comparison in the previous section, it is revealed that scene graph datasets are effective for TWC games because they have more spatial relationships. This suggests that the issue of the TWC agent is that the external knowledge is limited to being text-based. To address this, we propose an approach to use scene graph datasets as external knowledge to build a commonsense subgraph for agents.  Our proposal has two types depending on the training method and external knowledge.
\subsubsection{Scene Graph}
The simplest model is to replace the external knowledge of the TWC agent with scene graph datasets from ConceptNet. As shown in Table \ref{tab:comparison}, a scene graph dataset holds many triplets that are effective for solving TWC games, so we expect to obtain a higher score.
\subsubsection{ConceptNet + SG}
We also propose a method to complement the weak point of textual knowledge with scene graph datasets. Inspired by curricular learning~\cite{bengio2009curriculum}, we first provide the agent with textual knowledge and train it, then provide the same agent with external knowledge from scene graph datasets and continue training. We hypothesize that this method is effective in training agents that have commonsense knowledge balanced between abstract and concrete knowledge. In the first step, the overall commonsense is given by the textual knowledge. In the second step, the specific commonsense focused on location is given from the scene graph dataset.

\section{Experiments}
\subsection{Experimental Setup}
We conduct our experiments on TWC games~\cite{murugesan2021textbased}. A TWC game is a text-based game where the goal is to tidy up a house by putting objects where they should be. The connection between objects and the locations where they should be is not given by the game, so the agent needs to depend on commonsense knowledge. This domain has three difficulty levels (easy, medium, and hard) depending on the total number of objects in the game, the number of objects in which the agent needs to find their locations, and the number of rooms to explore. The numbers are randomly sampled from the list in Table~\ref{tab:twc_setting}. In our evaluation, we consider all difficulty levels. We also use two types of test sets: \textit{IN} and \textit{OUT}. The games in the \textit{IN} were built on the same entities as the training set, and the entities in the \textit{OUT} do not appear in the training set. We can evaluate the ability to generalize unseen entities from these test sets.

\begin{table}
   \centering
   \begin{tabular}[h]{>{\centering}m{0.2\linewidth}>{\raggedright}m{0.15\linewidth}>{\raggedright}m{0.25\linewidth}>{\raggedright\arraybackslash}m{0.15\linewidth}}
       \toprule
       Level & Objects & Objects to find & Rooms \\
       \midrule
       Easy & 1 & 1 & 1 \\
       Medium & 2, 3 & 1, 2, 3 & 1 \\
       Hard & 6, 7 & 5, 6, 7 & 1, 2 \\
       \bottomrule
   \end{tabular}
   \caption{Specification of TWC games from \cite{murugesan2021textbased}.}
   \label{tab:twc_setting}
\end{table}

Our experimental setup is based on the evaluation system in \cite{murugesan2021textbased}; we use the Advantage Actor-Critic algorithm~\cite{mnih2016asynchronous}. The most significant difference from the previous system is that all agents use GloVe for graph embedding. Numberbatch~\cite{numberbatch} used in the previous system is a combination of existing pre-trained embeddings such as word2vec~\cite{mikolov2013efficient} and GloVe retrofitted with ConceptNet's graph. The evaluation experiments in \cite{murugesan2021textbased} have shown that a TWC agent with Numberbatch achieved a better performance than with GloVe because of a high affinity with ConceptNet. Since we focus on the impact of external knowledge, we use only GloVe for both graph and observation embeddings in all agents.

\subsection{Metrics}
We measure the performance of agents with the various external knowledge on TextWorld using two metrics: the normalized score and the number of steps taken. The normalized score is calculated by dividing the actual score by the maximum possible score. Steps indicate time spent to reach the goal and the lower the value, the higher the performance.

\subsection{Results}

\begin{table*}
   \begin{threeparttable}[h]
      \small
       \centering
       \begin{tabular}[h]{>{\centering}m{0.01\linewidth}>{\centering}m{0.18\linewidth}>{\centering}m{0.105\linewidth}>{\centering}m{0.1\linewidth}>{\centering}m{0.105\linewidth}>{\centering}m{0.1\linewidth}>{\centering}m{0.105\linewidth}>{\centering\arraybackslash}m{0.1\linewidth}}
       \toprule
         & & \multicolumn{2}{c}{Easy} & \multicolumn{2}{c}{Medium} & \multicolumn{2}{c}{Hard} \\
        & Method & \#Steps$\downarrow$  & Score$\uparrow$ & \#Steps$\downarrow$ & Score$\uparrow$ & \#Steps$\downarrow$ & Score$\uparrow$ \\
       \midrule
       \multirow{4}{*}{\textit{IN}} & ConceptNet~\tnote{*} & $20.34 \pm 0.96$ & $0.82 \pm 0.04$ & $41.80 \pm 1.10$ & $0.65 \pm 0.06$ & $50.00 \pm 0.00$ & $ 0.25 \pm 0.07 $ \\
        & ConceptNet~+~Manual~\tnote{*} & $18.34 \pm 2.78$ & $0.85 \pm 0.05$ & \bm{$37.27 \pm 2.83$} & \bm{$0.77 \pm 0.06$} & \bm{$49.59 \pm 0.62$} & \bm{$0.35 \pm 0.03$} \\
        & \textbf{Scene Graph} & \bm{$17.23 \pm 3.16$} & \bm{$0.90 \pm 0.05$} & $42.18 \pm 2.66$ & $0.58 \pm 0.08$ & $50.00 \pm 0.00$ & $0.32 \pm 0.08$ \\
        & \textbf{ConceptNet + SG} & $20.43 \pm 1.28$ & $0.78 \pm 0.04$ & $38.30 \pm 6.08$ & $0.71 \pm 0.12$ & $50.00 \pm 0.00$ & $0.28 \pm 0.08$ \\
       \midrule
       \multirow{4}{*}{\textit{OUT}} & ConceptNet~\tnote{*} & $18.05 \pm 4.64$ & $0.88 \pm 0.08$ & $44.30 \pm 4.42$ & $0.50 \pm 0.09$ & \bm{$50.00 \pm 0.00$} & \bm{$0.19 \pm 0.04$} \\
        & ConceptNet~+~Manual~\tnote{*} & $27.01 \pm 2.72$ & $0.69 \pm 0.05$ & $46.44 \pm 1.15$ & $0.55 \pm 0.05$ & \bm{$50.00 \pm 0.00$} & \bm{$0.19 \pm 0.02$} \\
        & \textbf{Scene Graph} & \bm{$17.24 \pm 3.84$} & $0.91 \pm 0.05$ & \bm{$41.44 \pm 5.45$} & $0.55 \pm 0.14$ & \bm{$50.00 \pm 0.00$} & $0.15 \pm 0.06$  \\
        & \textbf{ConceptNet + SG} & $17.47 \pm 3.13$ & \bm{$0.92 \pm 0.04$} & $42.57 \pm 2.69$ & \bm{$0.63 \pm 0.04$} & \bm{$50.00 \pm 0.00$} & $0.13 \pm 0.05$ \\
       \bottomrule
       \end{tabular}
       \begin{tablenotes}
           \item[*] Baseline text-based RL agents with commonsense from external knowledge~\cite{murugesan2021textbased}
       \end{tablenotes}
   \end{threeparttable}
   \caption{Generalization results for two test sets, \textit{IN} and \textit{OUT}, on games with three difficulty levels. Scene Graph and ConceptNet~+~SG are our proposed methods, ConceptNet and ConceptNet~+~Manual are baseline agents with commonsense from external knowledge~\cite{murugesan2021textbased}. \textit{IN} is built using the same entities as the training set, and \textit{OUT} is built using different entities. \textbf{\#Steps} (lower is better) denotes the steps needed to accomplish the goals and \textbf{Score} (higher is better) denotes the normalized score by maximum possible score. Each value is a pair $(\textrm{average}) \pm (\textrm{standard deviation})$.}
   \label{tab:results}
\end{table*}

\begin{figure*}[h]
   \begin{center}
      \begin{tabular}{cc}
         \begin{minipage}{0.33\linewidth}
            \centering
            \includegraphics[width=\linewidth]{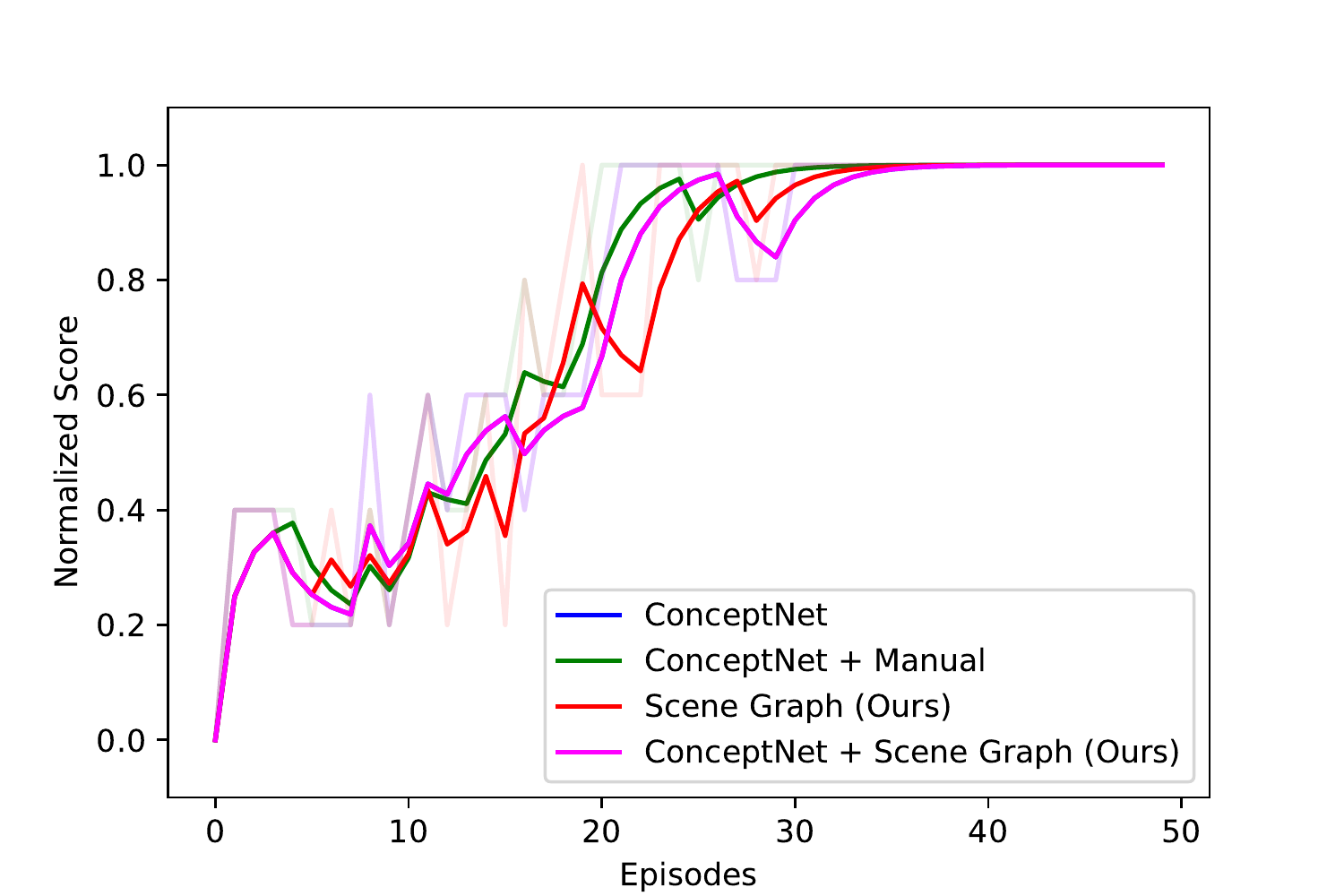}
            \subcaption{Easy, Normalized Score}
            \label{fig:easy_score}
         \end{minipage}
         \begin{minipage}{0.33\linewidth}
           \centering
           \includegraphics[width=\linewidth]{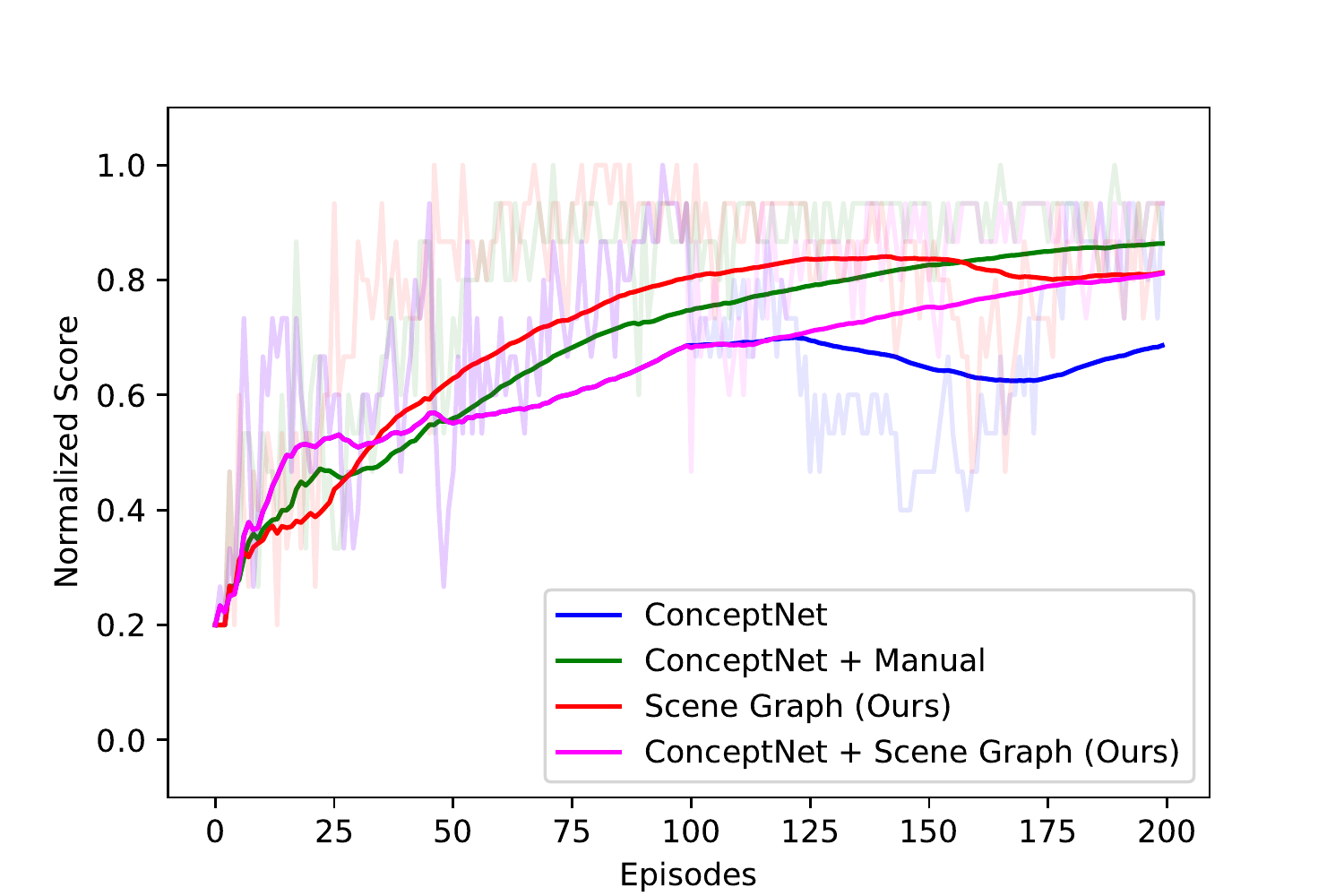}
           \subcaption{Medium, Normalized Score}
           \label{fig:medium_score}
        \end{minipage}
        \begin{minipage}{0.33\linewidth}
           \centering
           \includegraphics[width=\linewidth]{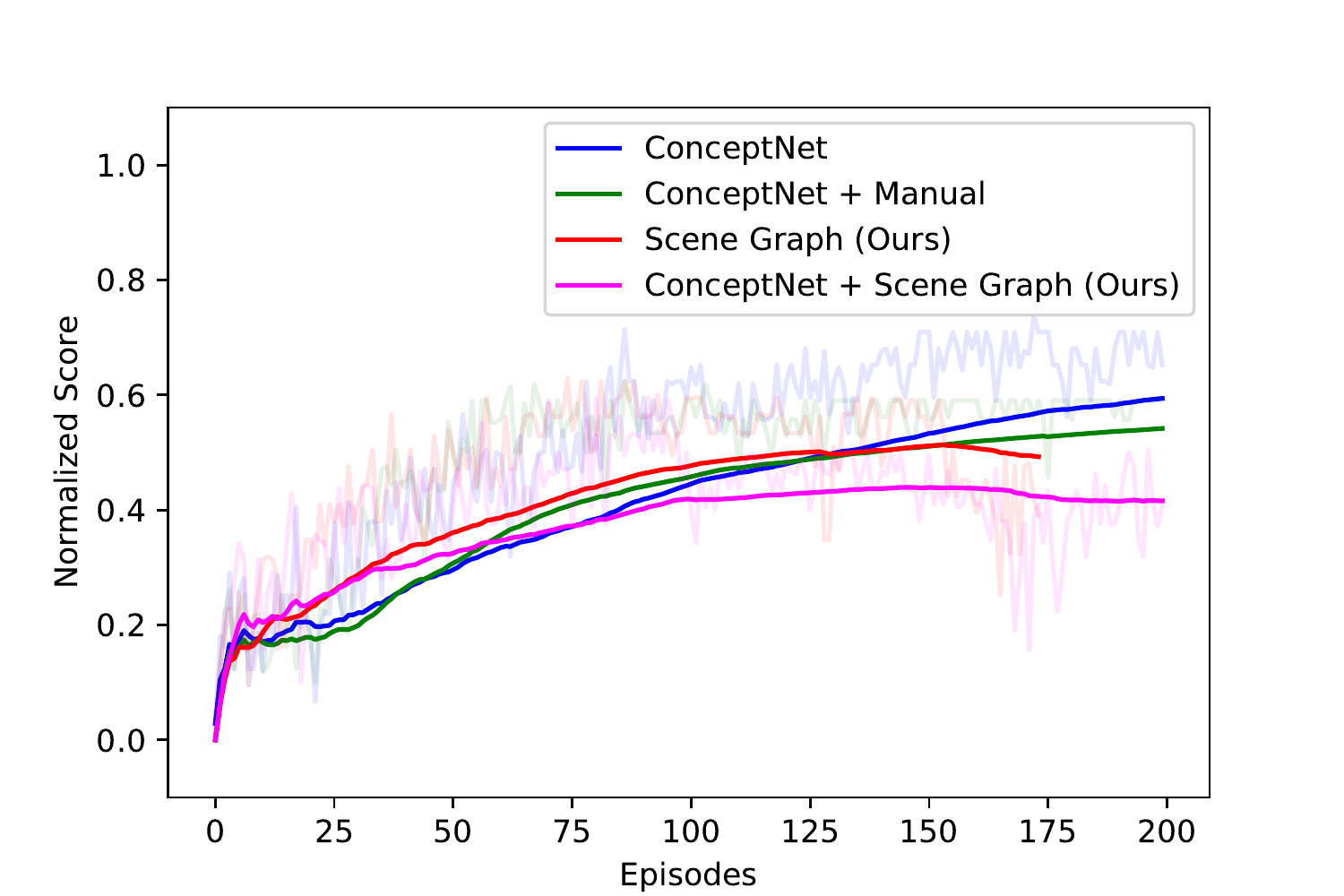}
           \subcaption{Hard, Normalized Score}
           \label{fig:hard_score}
        \end{minipage}
        \\
         \begin{minipage}{0.33\linewidth}
            \centering
            \includegraphics[width=\linewidth]{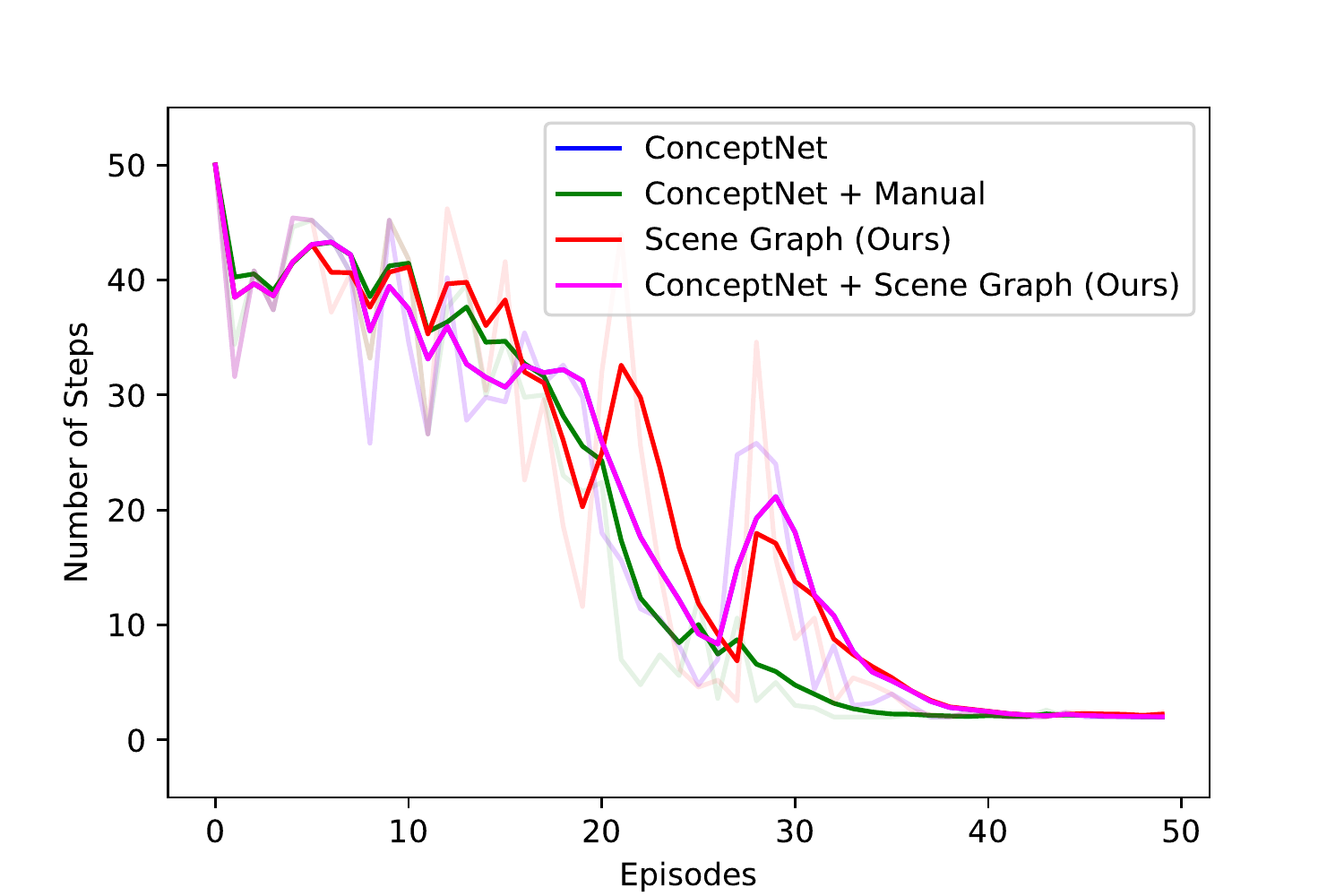}
            \subcaption{Easy, Steps}
            \label{fig:easy_steps}
         \end{minipage}
        \begin{minipage}{0.33\linewidth}
           \centering
           \includegraphics[width=\linewidth]{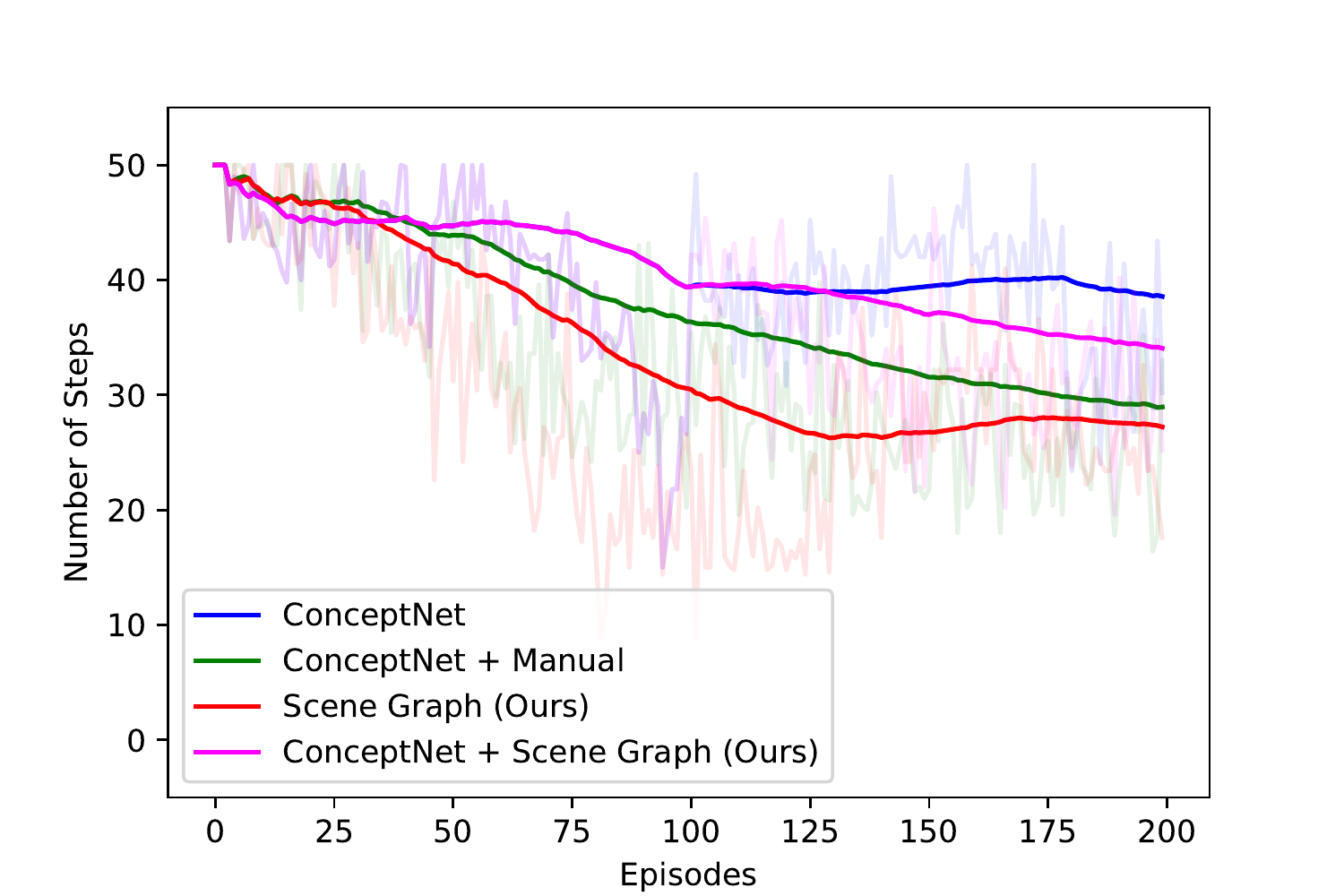}
           \subcaption{Medium, Steps}
           \label{fig:medium_steps}
        \end{minipage}
        \begin{minipage}{0.33\linewidth}
           \centering
           \includegraphics[width=\linewidth]{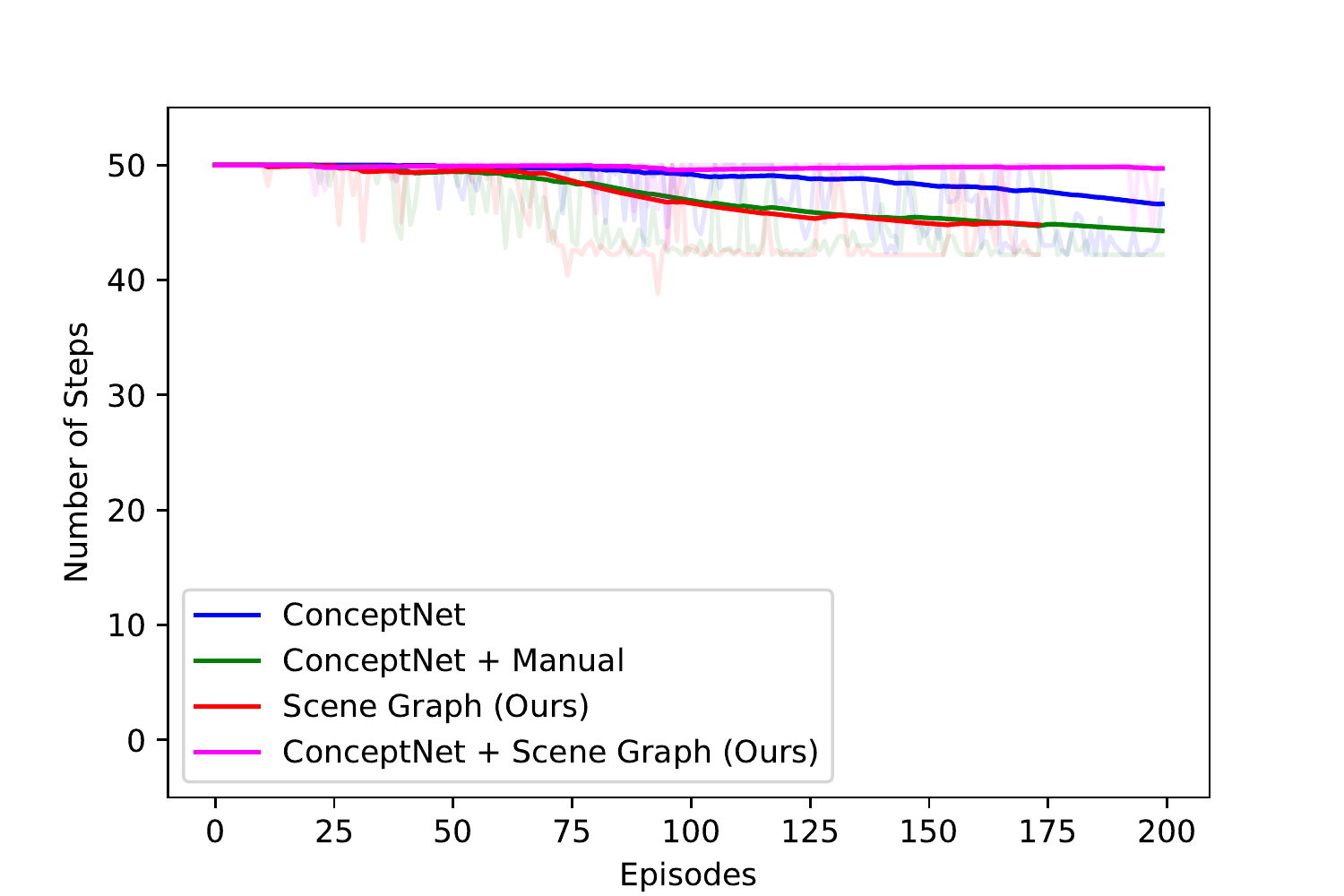}
           \subcaption{Hard, Steps}
           \label{fig:hard_steps}
        \end{minipage}
      \end{tabular}
      \caption{Performance evaluation for the three levels of the training set games (Smoothing is performed to clarify the difference in the results of a single run).}
      \label{fig:results}
   \end{center}
\end{figure*}

We compare four types of agents: ConceptNet, ConceptNet + Manual, Scene Graph, and ConceptNet + SG. Both Scene Graph and ConceptNet + SG are our proposed methods that use VG. The other agents are baselines proposed in \cite{murugesan2021textbased}. ConceptNet~+~Manual uses manually-prepared knowledge directly related to the game from ConceptNet. ConceptNet~+~Manual should show human-like performance and is regarded as the upper bound of the performance of the proposed method (especially for \text{IN}). ConceptNet~+~SG is first trained for 100 episodes using ConceptNet, followed by 100 episodes using VG. All agents except ConceptNet~+~SG are trained for 100 episodes each and all results are the average of five runs. The results are summarized in Table~\ref{tab:results}. We also show the training curves in Figure~\ref{fig:results}.

We can see three overall trends in these results. First, our proposed methods outperform the baselines in both steps and scores in the easy and medium levels. The commonsense knowledge obtained from scene graph datasets is proved to be effective in solving TWC games. Second, all agents struggle with the hard level. It is necessary to have the ability to deal with complicated situations where the location to pick up objects is different from the location to place them. The development of agents with this ability is future work. Finally, ConceptNet~+~Manual shows high performance for the \text{IN} set regardless of difficulty level. ConceptNet~+~Manual has crucial information for solving TWC games, so both efficiency and scores are higher when the same species of entity is given as the training set. However, its performance in the \textit{OUT} is lower than that of other agents because it is given only a minimum number of necessary graphs in the training set, so it overfits to those graphs.

We describe the performance of our proposed models in detail. We propose two models, Scene Graph and ConceptNet~+~SG, depending on the type of external knowledge and training method. 
Scene Graph is a simple agent that uses only a scene graph dataset. From Table~\ref{tab:results}, we found that this agent is very efficient in its graph search. For easy-level games, Scene Graph is superior to even ConceptNet~+~Manual in the \textit{IN} set. In medium-level games, Scene Graph has the largest number of steps in the \textit{IN} set, but it has the smallest number in the \text{OUT} set. This indicates that Scene Graph is robust against unseen objects. The efficiency can be seen from the fastest convergence of the training curves in the medium and hard levels in Figure~\ref{fig:results}. In terms of performance, although it is sometimes inferior to our other proposed method, it is better than the baselines in \textit{OUT}. The high efficiency and performance of Scene Graph can be attributed to scene graph datasets having many graphs relevant to the TWC game with spatial relationships. 
ConceptNet~+~SG is an agent that is trained with ConceptNet, followed by scene graph datasets. Table~\ref{tab:results} shows that the performance of this agent is very high. It has the best performance in the \textit{OUT} set for both easy and medium levels and the second-best performance in the \textit{IN} set after ConceptNet~+~Manual. This indicates that it also has robustness against unseen objects. The reason for the performance improvement is considered to be the wide range that can be handled by both commonsense knowledge from ConceptNet and VG. In easy-level games, the score increases by 0.14 from \textit{IN} to \textit{OUT}, which is an uncommon improvement. The reason for this could be that VG has a small number of graphs related to the easy-\textit{IN} games, which decrease the results of ConceptNet~+~SG for easy-\text{IN} games. As you can see in Table~\ref{tab:twc_setting}, easy games has only require one object to be explored, so the score changes dramatically depending on whether or not external knowledge contains graphs related to the object and the goal location. 
However, since the number of graphs increases, the efficiency of the search decreases, resulting in inferiority with scene graphs in steps. The training curve in Figure~\ref{fig:results} shows that VG enhances the performance of ConceptNet alone from 100 episodes (in the easy level, the training curve converges during the ConceptNet phase, so ConceptNet and ConceptNet~+~SG overlap).

In addition, we use GloVe for graph embedding in these experiments for a fair comparison, but agents using ConceptNet can be improved by replacing GloVe with  Numberbatch.

In summary, our proposed method achieves both efficiency and performance improvements, and it is robust to unseen objects. However, it is still a challenge to deal with complex situations such as hard-level games. We discuss how to address this issue in the next section.

\section{Conclusion and Future Work}
We have presented new approaches to leverage commonsense subgraphs constructed from scene graph datasets for text-based games. We conducted experiments on a TWC game, which is a benchmark to evaluate how well an agent learns with commonsense knowledge. Experimental results showed that our proposed approaches using a VG dataset demonstrate highly competitive performances compared with existing state-of-the-art approaches with textual knowledge. We also illustrated that the performance can be further improved by using ConceptNet and scene graph datasets sequentially.

Although this work is the first step to utilize both visual information and commonsense, we still have a few challenges as future work. One topic is to deal with more complex situations like hard difficulty level games. We expect that this can be improved by exploiting the relationships among information available from external knowledge. The current model adopts GAT~\cite{velickovic2018graph} as a graph encoder, but GAT only takes the node features as input and ignores the edge features except for whether they exist or not. In complex tasks, the key to solving this game is more specific information than simply the link between an object and location. In particular, scene graph datasets provide more detailed information on relationships than textual knowledge as shown in Fig.~\ref{fig:teaser}. We consider applying networks such as Edge feature enhanced Graph Neural Networks (EGNN)~\cite{gong2019exploiting} that can take advantage of the edge features of graphs. Another topic is introducing logical rule training into our proposed method. Since graph information can be easily converted to logical rules, we hope the commonsense graph can directly contribute to logical rule training for action policies in RL.

\bibliography{aaai22}

\begin{thebibliography}{32}
\providecommand{\natexlab}[1]{#1}

\bibitem[{Ammanabrolu and Riedl(2021)}]{ammanabrolu2021learning}
Ammanabrolu, P.; and Riedl, M.~O. 2021.
\newblock Learning Knowledge Graph-based World Models of Textual Environments.
\newblock \emph{arXiv preprint arXiv:2106.09608}.

\bibitem[{Bengio et~al.(2009)Bengio, Louradour, Collobert, and
  Weston}]{bengio2009curriculum}
Bengio, Y.; Louradour, J.; Collobert, R.; and Weston, J. 2009.
\newblock Curriculum learning.
\newblock In \emph{Proceedings of the 26th annual international conference on
  machine learning}, 41--48.

\bibitem[{Carta et~al.(2020)Carta, Chaudhury, Talamadupula, and
  Tatsubori}]{carta-etal-2020-vizhints}
Carta, T.; Chaudhury, S.; Talamadupula, K.; and Tatsubori, M. 2020.
\newblock VisualHints: A Visual-Lingual Environment for Multimodal
  Reinforcement Learning.
\newblock In \emph{arxiv}.

\bibitem[{Chaudhury et~al.(2021)Chaudhury, Sen, Ono, Kimura, Tatsubori, and
  Munawar}]{chaudhury-etal-2021-neuro}
Chaudhury, S.; Sen, P.; Ono, M.; Kimura, D.; Tatsubori, M.; and Munawar, A.
  2021.
\newblock Neuro-Symbolic Approaches for Text-Based Policy Learning.
\newblock In \emph{Proceedings of the 2021 Conference on Empirical Methods in
  Natural Language Processing}, 3073--3078. Online and Punta Cana, Dominican
  Republic: Association for Computational Linguistics.

\bibitem[{C\^ot\'e et~al.(2018)C\^ot\'e, K\'ad\'ar, Yuan, Kybartas, Barnes,
  Fine, Moore, Tao, Hausknecht, Asri, Adada, Tay, and
  Trischler}]{cote18textworld}
C\^ot\'e, M.-A.; K\'ad\'ar, A.; Yuan, X.; Kybartas, B.; Barnes, T.; Fine, E.;
  Moore, J.; Tao, R.~Y.; Hausknecht, M.; Asri, L.~E.; Adada, M.; Tay, W.; and
  Trischler, A. 2018.
\newblock TextWorld: A Learning Environment for Text-based Games.
\newblock \emph{CoRR}, abs/1806.11532.

\bibitem[{Gong and Cheng(2019)}]{gong2019exploiting}
Gong, L.; and Cheng, Q. 2019.
\newblock Exploiting edge features for graph neural networks.
\newblock In \emph{Proceedings of the IEEE/CVF Conference on Computer Vision
  and Pattern Recognition}, 9211--9219.

\bibitem[{Hausknecht et~al.(2019)Hausknecht, Ammanabrolu, C^^c3^^b4t^^c3^^a9,
  and Yuan}]{hausknecht2019interactive}
Hausknecht, M.; Ammanabrolu, P.; C^^c3^^b4t^^c3^^a9, M.-A.; and Yuan, X.~E.
  2019.
\newblock Interactive Fiction Games: A Colossal Adventure.
\newblock In \emph{AAAI 2020}.

\bibitem[{Kaelbling, Littman, and Cassandra(1998)}]{pomdp}
Kaelbling, L.~P.; Littman, M.~L.; and Cassandra, A.~R. 1998.
\newblock Planning and acting in partially observable stochastic domains.
\newblock \emph{Artificial Intelligence}, 101(1): 99--134.

\bibitem[{Kimura(2018)}]{daqn}
Kimura, D. 2018.
\newblock DAQN: Deep Auto-encoder and Q-Network.
\newblock \emph{arXiv:1806.00630}.

\bibitem[{Kimura et~al.(2020)Kimura, Chaudhury, Narita, Munawar, and
  Tachibana}]{9093428}
Kimura, D.; Chaudhury, S.; Narita, M.; Munawar, A.; and Tachibana, R. 2020.
\newblock Adversarial Discriminative Attention for Robust Anomaly Detection.
\newblock In \emph{2020 IEEE Winter Conference on Applications of Computer
  Vision (WACV)}, 2161--2170.

\bibitem[{Kimura et~al.(2021{\natexlab{a}})Kimura, Chaudhury, Ono, Tatsubori,
  Agravante, Munawar, Wachi, Kohita, and Gray}]{kimura-etal-2021-loa}
Kimura, D.; Chaudhury, S.; Ono, M.; Tatsubori, M.; Agravante, D.~J.; Munawar,
  A.; Wachi, A.; Kohita, R.; and Gray, A. 2021{\natexlab{a}}.
\newblock {LOA}: Logical Optimal Actions for Text-based Interaction Games.
\newblock In \emph{Proceedings of the 59th Annual Meeting of the Association
  for Computational Linguistics and the 11th International Joint Conference on
  Natural Language Processing: System Demonstrations}, 227--231. Online:
  Association for Computational Linguistics.

\bibitem[{Kimura et~al.(2018)Kimura, Chaudhury, Tachibana, and
  Dasgupta}]{marioirl}
Kimura, D.; Chaudhury, S.; Tachibana, R.; and Dasgupta, S. 2018.
\newblock Internal Model from Observations for Reward Shaping.
\newblock In \emph{ICML workshop}.

\bibitem[{Kimura et~al.(2021{\natexlab{b}})Kimura, Chaudhury, Wachi, Kohita,
  Munawar, Tatsubori, and Gray}]{kimura2020reinforcement}
Kimura, D.; Chaudhury, S.; Wachi, A.; Kohita, R.; Munawar, A.; Tatsubori, M.;
  and Gray, A. 2021{\natexlab{b}}.
\newblock Reinforcement Learning with External Knowledge by using Logical
  Neural Networks.
\newblock \emph{KBRL Workshop at IJCAI-PRICAI 2020}.

\bibitem[{Kimura et~al.(2021{\natexlab{c}})Kimura, Ono, Chaudhury, Kohita,
  Wachi, Agravante, Tatsubori, Munawar, and Gray}]{kimura-etal-2021-neuro}
Kimura, D.; Ono, M.; Chaudhury, S.; Kohita, R.; Wachi, A.; Agravante, D.~J.;
  Tatsubori, M.; Munawar, A.; and Gray, A. 2021{\natexlab{c}}.
\newblock Neuro-Symbolic Reinforcement Learning with First-Order Logic.
\newblock In \emph{Proceedings of the 2021 Conference on Empirical Methods in
  Natural Language Processing}, 3505--3511. Online and Punta Cana, Dominican
  Republic: Association for Computational Linguistics.

\bibitem[{Krishna et~al.(2017)Krishna, Zhu, Groth, Johnson, Hata, Kravitz,
  Chen, Kalantidis, Li, Shamma et~al.}]{krishna2017visual}
Krishna, R.; Zhu, Y.; Groth, O.; Johnson, J.; Hata, K.; Kravitz, J.; Chen, S.;
  Kalantidis, Y.; Li, L.-J.; Shamma, D.~A.; et~al. 2017.
\newblock Visual genome: Connecting language and vision using crowdsourced
  dense image annotations.
\newblock \emph{International journal of computer vision}, 123(1): 32--73.

\bibitem[{Liu and Singh(2004)}]{Liu2004ConceptNetA}
Liu, H.; and Singh, P. 2004.
\newblock ConceptNet ― A Practical Commonsense Reasoning Tool-Kit.
\newblock \emph{BT Technology Journal}, 22: 211--226.

\bibitem[{Lu et~al.(2017)Lu, Xiong, Parikh, and Socher}]{lu2017knowing}
Lu, J.; Xiong, C.; Parikh, D.; and Socher, R. 2017.
\newblock Knowing when to look: Adaptive attention via a visual sentinel for
  image captioning.
\newblock In \emph{Proceedings of the IEEE conference on computer vision and
  pattern recognition}, 375--383.

\bibitem[{Mikolov et~al.(2013)Mikolov, Chen, Corrado, and
  Dean}]{mikolov2013efficient}
Mikolov, T.; Chen, K.; Corrado, G.; and Dean, J. 2013.
\newblock Efficient estimation of word representations in vector space.
\newblock \emph{arXiv preprint arXiv:1301.3781}.

\bibitem[{Mnih et~al.(2016)Mnih, Badia, Mirza, Graves, Lillicrap, Harley,
  Silver, and Kavukcuoglu}]{mnih2016asynchronous}
Mnih, V.; Badia, A.~P.; Mirza, M.; Graves, A.; Lillicrap, T.; Harley, T.;
  Silver, D.; and Kavukcuoglu, K. 2016.
\newblock Asynchronous methods for deep reinforcement learning.
\newblock In \emph{International conference on machine learning}, 1928--1937.
  PMLR.

\bibitem[{Mnih et~al.(2015)Mnih, Kavukcuoglu, Silver, and et~al.}]{dqn}
Mnih, V.; Kavukcuoglu, K.; Silver, D.; and et~al. 2015.
\newblock Human-level control through deep reinforcement learning.
\newblock \emph{Nature}.

\bibitem[{Murugesan et~al.(2021)Murugesan, Atzeni, Kapanipathi, Shukla,
  Kumaravel, Tesauro, Talamadupula, Sachan, and
  Campbell}]{murugesan2021textbased}
Murugesan, K.; Atzeni, M.; Kapanipathi, P.; Shukla, P.; Kumaravel, S.; Tesauro,
  G.; Talamadupula, K.; Sachan, M.; and Campbell, M. 2021.
\newblock Text-based RL Agents with Commonsense Knowledge: New Challenges,
  Environments and Baselines.
\newblock In \emph{Thirty Fifth AAAI Conference on Artificial Intelligence}.

\bibitem[{Murugesan, Chaudhury, and Talamadupula(2021)}]{murugesan2021eye}
Murugesan, K.; Chaudhury, S.; and Talamadupula, K. 2021.
\newblock Eye of the Beholder: Improved Relation Generalization for Text-based
  Reinforcement Learning Agents.
\newblock \emph{arXiv preprint arXiv:2106.05387}.

\bibitem[{Pennington, Socher, and Manning(2014)}]{pennington-etal-2014-glove}
Pennington, J.; Socher, R.; and Manning, C. 2014.
\newblock {G}lo{V}e: Global Vectors for Word Representation.
\newblock In \emph{Proceedings of the 2014 Conference on Empirical Methods in
  Natural Language Processing ({EMNLP})}, 1532--1543. Doha, Qatar: Association
  for Computational Linguistics.

\bibitem[{Riegel et~al.(2020)Riegel, Gray, Luus, Khan, Makondo, Akhalwaya,
  Qian, Fagin, Barahona, Sharma et~al.}]{riegel2020logical}
Riegel, R.; Gray, A.; Luus, F.; Khan, N.; Makondo, N.; Akhalwaya, I.~Y.; Qian,
  H.; Fagin, R.; Barahona, F.; Sharma, U.; et~al. 2020.
\newblock Logical neural networks.
\newblock \emph{arXiv preprint arXiv:2006.13155}.

\bibitem[{Shridhar et~al.(2020)Shridhar, Thomason, Gordon, Bisk, Han, Mottaghi,
  Zettlemoyer, and Fox}]{ALFRED20}
Shridhar, M.; Thomason, J.; Gordon, D.; Bisk, Y.; Han, W.; Mottaghi, R.;
  Zettlemoyer, L.; and Fox, D. 2020.
\newblock {ALFRED: A Benchmark for Interpreting Grounded Instructions for
  Everyday Tasks}.
\newblock In \emph{The IEEE Conference on Computer Vision and Pattern
  Recognition (CVPR)}.

\bibitem[{Shridhar et~al.(2021)Shridhar, Yuan, C\^ot\'e, Bisk, Trischler, and
  Hausknecht}]{ALFWorld20}
Shridhar, M.; Yuan, X.; C\^ot\'e, M.-A.; Bisk, Y.; Trischler, A.; and
  Hausknecht, M. 2021.
\newblock {ALFWorld: Aligning Text and Embodied Environments for Interactive
  Learning}.
\newblock In \emph{Proceedings of the International Conference on Learning
  Representations (ICLR)}.

\bibitem[{Silver, Huang, and et~al.(2016)}]{alphago}
Silver, D.; Huang, A.; and et~al. 2016.
\newblock Mastering the game of Go with deep neural networks and tree search.
\newblock \emph{Nature}, 529: 484--503.

\bibitem[{Speer, Chin, and Havasi(2017{\natexlab{a}})}]{ConceptNet2017}
Speer, R.; Chin, J.; and Havasi, C. 2017{\natexlab{a}}.
\newblock ConceptNet 5.5: An Open Multilingual Graph of General Knowledge.
\newblock In \emph{Proceedings of the Thirty-First AAAI Conference on
  Artificial Intelligence}, AAAI'17, 4444^^e2^^80^^934451. AAAI Press.

\bibitem[{Speer, Chin, and Havasi(2017{\natexlab{b}})}]{numberbatch}
Speer, R.; Chin, J.; and Havasi, C. 2017{\natexlab{b}}.
\newblock ConceptNet 5.5: An Open Multilingual Graph of General Knowledge.
\newblock AAAI'17, 4444^^e2^^80^^934451. AAAI Press.

\bibitem[{Tanaka and Simo-Serra(2021)}]{Tanaka_2021_CVPR}
Tanaka, T.; and Simo-Serra, E. 2021.
\newblock LoL-V2T: Large-Scale Esports Video Description Dataset.
\newblock In \emph{Proceedings of the IEEE/CVF Conference on Computer Vision
  and Pattern Recognition (CVPR) Workshops}, 4557--4566.

\bibitem[{Veli{\v{c}}kovi{\'{c}} et~al.(2018)Veli{\v{c}}kovi{\'{c}}, Cucurull,
  Casanova, Romero, Li{\`{o}}, and Bengio}]{velickovic2018graph}
Veli{\v{c}}kovi{\'{c}}, P.; Cucurull, G.; Casanova, A.; Romero, A.; Li{\`{o}},
  P.; and Bengio, Y. 2018.
\newblock {Graph Attention Networks}.
\newblock \emph{International Conference on Learning Representations}.

\bibitem[{Xu et~al.(2018)Xu, Zhang, Huang, Zhang, Gan, Huang, and
  He}]{Tao18attngan}
Xu, T.; Zhang, P.; Huang, Q.; Zhang, H.; Gan, Z.; Huang, X.; and He, X. 2018.
\newblock AttnGAN: Fine-Grained Text to Image Generation with Attentional
  Generative Adversarial Networks.

\end{thebibliography}

\end{document}